\documentclass[10pt,twocolumn,letterpaper]{article}

\usepackage{cvpr}
\usepackage{times}
\usepackage{epsfig}
\usepackage{graphicx}
\usepackage{subfig}
\usepackage{amsmath}
\usepackage{amssymb}
\usepackage{bbm}
\usepackage{algorithm}
\usepackage[noend]{algpseudocode}
\usepackage{algpseudocode}
\usepackage{authblk}
\makeatletter
\def\BState{\State\hskip-\ALG@thistlm}
\makeatother

\usepackage[pagebackref=true,breaklinks=true,letterpaper=true,colorlinks,bookmarks=false]{hyperref}
\cvprfinalcopy % *** Uncomment this line for the final submission

\def\cvprPaperID{345} % *** Enter the CVPR Paper ID here

\ifcvprfinal\fi
\begin{document}
%%%%%%%%% TITLE
\title{4D Human Body Correspondences from Panoramic Depth Maps}
%\title{Correspondence Matching and Compression on 4D Human Avatars Using Panoramic Depth Maps}

%\author{Yingliang Zhang$^{\text{1}}$\qquad Peihong Yu$^{\text{1}}$ \qquad  Wei Yang$^{\text{2,3}}$ \qquad  Yuanxi Ma$^{\text{1}}$ \qquad Jingyi Yu$^{\text{1,3}}$ \\
%\and
%$^{\text{1}}$ShanghaiTech University  \\
%{\tt \small\{zhangyl,yuph,mayx\}@shanghaitech.edu.cn}
%\and
%$^{\text{2}}$University of Delaware  \\
%{\tt \small wyangcs@udel.edu}
%\and
%$^{\text{3}}$Plex-VR Inc.  \\
%{\tt \small jingyi.yu@plex-vr.com}
%}
%\maketitle

%\author{Zhong Li$^{\text{1}}$ \qquad Minye Wu$^{\text{2}}$ \qquad  Yitengwang Zhou$^{\text{2}}$ \qquad Jingyi Yu$^{\text{2}}$ \\
%\and
%$^{\text{1}}$University of Delaware,Newark,USA  \\
%{\tt \small lizhong@udel.edu}
%\and
%$^{\text{2}}$ShanghaiTech University,Shanghai,China  \\
%{\tt \small \{wumy,yujy1g\}@shanghaitech.edu}
%%}

\newcommand\CoAuthorMark{\footnotemark[\arabic{footnote}]}
\newcommand*\samethanks[1][\value{footnote}]{\footnotemark[#1]}
\renewcommand\Authands{ and }

\author[1]{Zhong Li\thanks{These authors contribute to the work equally.}}
\author[2]{Minye Wu\samethanks}
\author[2]{Wangyiteng Zhou}
\author[1,2]{Jingyi Yu}
\affil[1]{University of Delaware, Newark, DE, USA. \texttt{lizhong@udel.edu}}
\affil[2]{ShanghaiTech University, Shanghai, China. \texttt{\{wumy,yujingyi\}@shanghaitech.edu.cn} \texttt{wytzhou@gmail.com}}

\maketitle
%\thispagestyle{empty}

%%%%%%%%% ABSTRACT
\begin{abstract}
   The availability of affordable 3D full body reconstruction systems has given rise to free-viewpoint video (FVV) of human shapes. Most existing solutions produce temporally uncorrelated point clouds or meshes with unknown point/vertex correspondences. Individually compressing each frame is ineffective and still yields to ultra-large data sizes. We present an end-to-end deep learning scheme to establish dense shape correspondences and subsequently compress the data. Our approach uses sparse set of "panoramic" depth maps or PDMs, each emulating an inward-viewing concentric mosaics (CM)~\cite{Shum1999RenderingWC}. We then develop a learning-based technique to learn pixel-wise feature descriptors on PDMs. The results are fed into an autoencoder-based network for compression. Comprehensive experiments demonstrate our solution is robust and effective on both public and our newly captured datasets.
\end{abstract}

%%%%%%%%% BODY TEXT
\section{Introduction}
There is an emerging trend on producing free-viewpoint video (FVV) of dynamic 3D human models~\cite{Collet2015HighqualitySF}, to provide viewers an unprecedented immersive viewing experience. The technology is largely enabled by the availability of affordable 3D acquisition systems and reliable reconstruction algorithms. The earlier attempt of Kanade et al~\cite{Kanade2007VirtualizedRP} mounted 51 cameras on a 5 meter diameter dome to "virtualize" reality. More recent solutions can be viewed as it variations but using higher resolution, higher speed industrial cameras and easy-to-use synchronization schemes. For example, the CMU Panoptic studio~\cite{Joo2015PanopticSA} uses 480 cameras and can recover interactions between multiple human subjects. Active solutions such Microsoft Holoportation~\cite{Orts2016HoloportationV3} further employ structured light to reduce the number of cameras.

\begin{figure}
\centering
\includegraphics[width=0.90\linewidth]{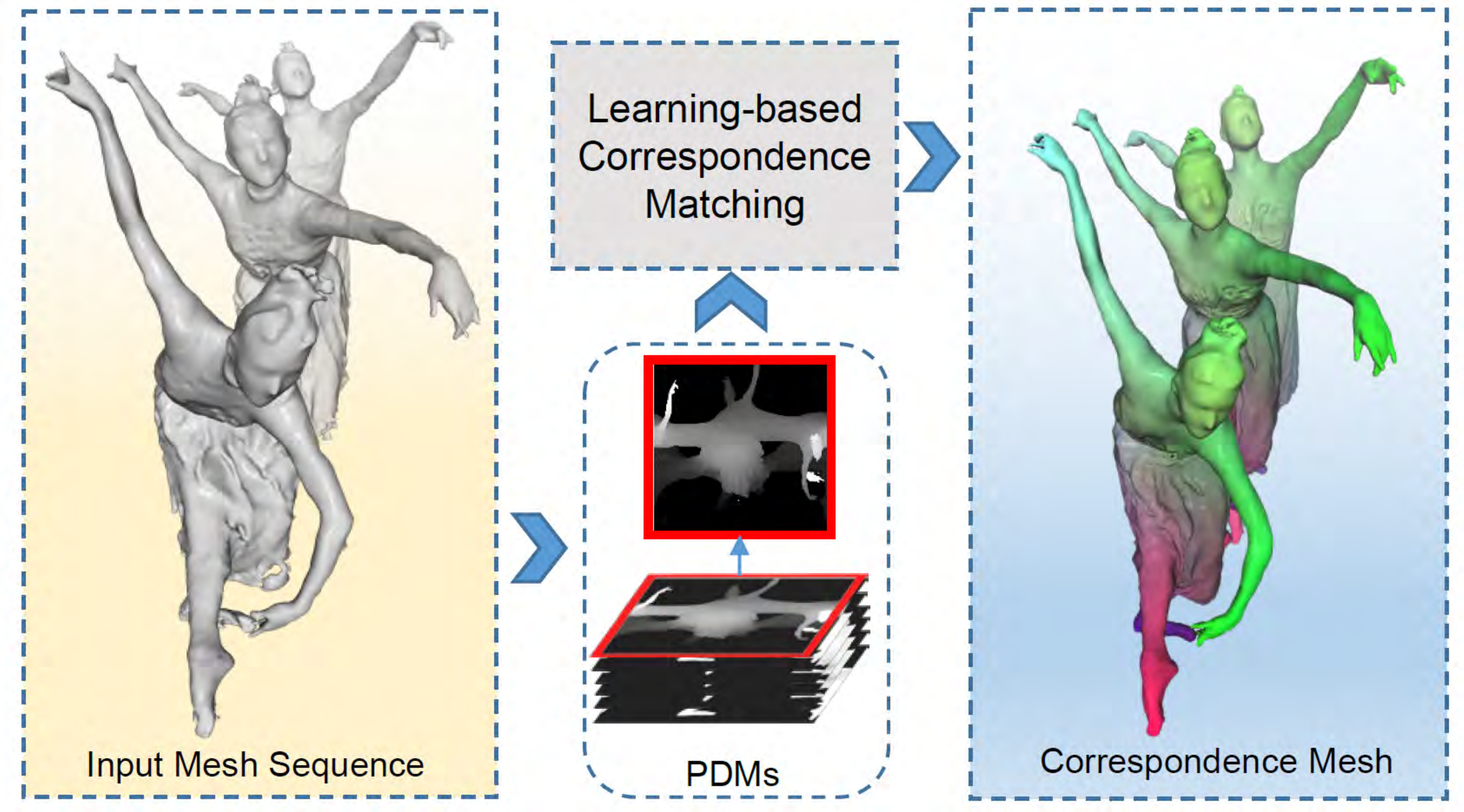}
\caption{Our human body correspondence technique first renders Panoramic Depth Maps (PDMs) of the input mesh sequences and then conduct learning-based correspondence matching on the PDMs.}
\label{fig:teaser}
\end{figure}

Despite heterogeneity in the digitization processes, a common challenge in FVV is the size of the reconstructed 4D data: each frame corresponds to a dense 3d mesh and a high resolution texture map and a short clip can easily lead to gigabytes of data if not compressed. For example, a sample clip of 10 seconds released by 8i~\cite{8i} is 2 gigabytes. The large data size prohibits real-time transfer to or even storage on user-end devices. Although existing video compression standards can compress the texture maps and potentially the mesh, they ignore geometric consistencies and yields to low compression rate and quality.

%a startup company focusing on dome based human reconstruction,

\begin{figure*}
\centering
\includegraphics[width=0.85\linewidth]{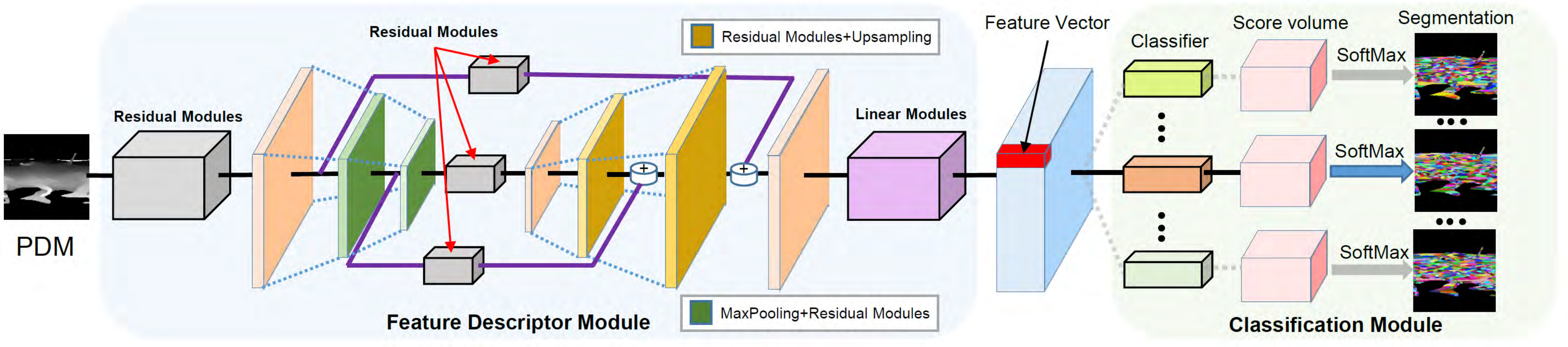}
\caption{Our human body correspondences network structure extends the hourglass network. The feature descriptor module learns per-pixel feature vectors on the PDMs where the results, along with body segmentations, are fed into the classification module.}
\label{fig:CorresNetworkPipeline}
\end{figure*}

The key to geometry-consistent compression is reliably establishing correspondences between geometric shapes. On 4D human body (geometry + time), the task is particularly challenging as such scans exhibit high noise, large non-rigid deformations, and topology changes. \textcolor{black}{Existing approaches assume small deformations so that sparse shape descriptors~\cite{Rusinkiewicz20053D} can be adopted. In reality, sparse shape descriptors fail on noisy data. Alternative dense shape correspondence schemes can reliably handle noise but require zero-genus surfaces}, i.e., they are inapplicable to topology changes. A notable exception is the recent deep-learning based approach~\cite{wei2016dense} that first trains a feature descriptor on depth maps produced from a large number of viewpoints {\color{black} 144 views} to classify the body regions.

\begin{figure}
\centering
\includegraphics[width=0.9\linewidth]{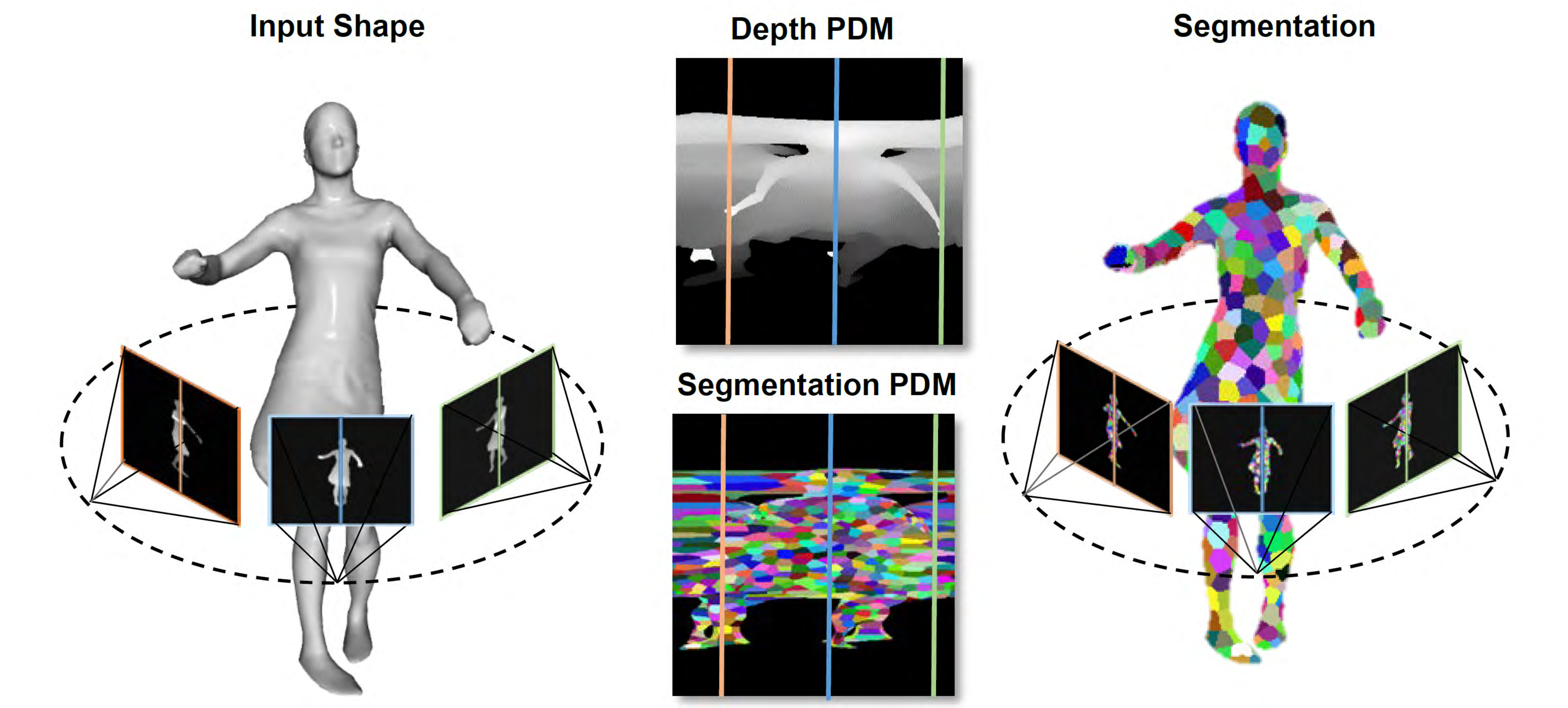}
\caption{A PDM represents an omni-directional depth map of a 3D mesh. We can also generate the corresponding PDM segmentation map (color coded).}
\label{fig:PDM_fig} %temporal_hint
\end{figure}

In a similar vein, we present an end-to-end deep learning scheme to conduct dense shape correspondences (as shown in Fig.~\ref{fig:teaser})  and subsequently compress the data. The key difference is that we aim to directly handle the complete 3D model without sampling depth maps from dense viewpoints. At each frame, we first produce a sparse set of "panoramic" depth maps or PDMs of 3D human model. Specifically, we construct 6 inward-viewing concentric mosaics (CM)~\cite{Shum1999RenderingWC} towards each model as shown in Fig.~\ref{fig:PDM_fig}. Tradition CMs are synthesized by rendering an outward viewing cameras that lie on a common circle and then stitching the central column of each image into a panorama. The collected rays are multi-perspective and hence ensure the sampling of visible and occluded regions.

To conduct efficient training, we apply GPU-based multi-perspective rendering~\cite{Yu2009TowardsMR} of PDMs for a variety of 3D human sequences (the MIT dataset~\cite{vlasic2009dynamic}, SCAPE~\cite{Anguelov2005SCAPESC}, and Yobi3D~\cite{wei2016dense}). Next, we extend the hourglass networks~\cite{newell2016stacked} to learn a pixel-wise feature descriptor for distinguishing different body parts on PDMs. We further add a regularization term in the network loss function to maximize the distance between feature descriptors that belong to different body parts. Once we obtain the pixel-wise feature descriptor on each pixel of the PMDs, we back-project it onto the 3D models for computing vertex-wise feature descriptors.

Since each vertex can be potentially mapped to a pixel in each of the 6 PDMs, we set out to find the most consistent matching vertex pairs across all 6 PDMs via a voting scheme. To further remove the outliers, we conduct correspondence fine-tuning based on the observation that the trajectory of each vertex induced by matching should be temporally smooth. We also improve the accuracy of correspondences by enforcing the geodesic constraints~\cite{huang2008non}. The process can significantly reduce the outliers while maintaining smooth motion trajectories. Finally, we feed the correspondences to an autoencoder-based network for geometry compression.

We conduct comprehensive experiments on a wide range of existing and our own 3D human motion sequences. {\color{black} Compared with ~\cite{wei2016dense}, the use of the PDMs significantly reduces the training data sizes. More importantly, PDMs are able to robustly handle occlusions in body geometry, yielding to a more reliable feature descriptor for correspondence matching. On the FAUST benchmark~\cite{bogo2014faust} our technique outperforms the state-of-the-art techniques and at the same time avoids complex optimizations. Further, our neural network based compression/decompression scheme achieves very high compression rate with low loss on both public and our newly captured datasets. }

 %To compute dense correspondence between two triangle mesh
%------------------------------------------------------------------------
\section{Related Work}
The key to any successful dynamic mesh compression scheme is to establish accurate vertex correspondences. Lipman et al.~\cite{lipman2009mobius} and Kim et al.~\cite{kim2010mobius} employed conformal geometric constraint between two frames. Such techniques are computationally expensive and require topological consistency. Bronstein et al. conduct vertex correspondence matching by imposing geodesic ~\cite{bronstein2010gromov} or diffusion ~\cite{bronstein2006generalized} distance constraints. Their techniques, however, still assume that input surfaces are nearly isometric and therefore cannot handle complex, articulated human motions. More recent approaches aim to design feature descriptors ~\cite{windheuser2014optimal,litman2014learning} matching points.
Pottemann et al.~\cite{pottmann2009integral} use local geometric descriptors that can handle small motions. Taylor et al.~\cite{Taylor2012TheVM} use random decision forest based approaches to infer correspondences.

Bogo et al.~\cite{bogo2014faust,Bogo2017DynamicFR} build a high-quality inter-shape correspondence benchmark by painting the human subject with high-frequency textures. Chen et al.~\cite{chen2015robust} manage to use the Markov Random Field (MRF) to solve correspondence matching analogous to \textcolor{black}{stereo matching}. Yet their technique is vulnerable to human wearing cloths. Most recently learning-based approaches such as the ones based on anisotropic convolutional neural network ~\cite{boscaini2016learning} have shown promising results on mesh correspondence matching. Yet the state-of-the-art solution~\cite{wei2016dense} requires sampling the mesh from a large number of viewpoints (144 in their solution) to reliably learn per-pixel feature descriptor. In contrast, we show how to train a network using a few of panoramic images. Further, the focus of their work is on \textcolor{black}{dense correspondence} rather than compression as ours where we show the latter requires higher temporal coherence.

Different from mesh correspondence matching, animated mesh compression is a well-studied problem in computer graphics. State-of-the-art solutions however assume consecutive meshes have exact connectivity ~\cite{maglo20153d}. \cite{gupta2002compression,mamou2009tfan} conduct pre-segmentation on the mesh to ensure connectivity consistencies.
PCA-based methods~\cite{Alexa2000RepresentingAB,Vasa2007CODDYAC,sattler2005simple,Luo2013CompressionO3} aim to identify different geometric clusters of human body (arms, hands, legs, torso, head, etc). Spatio-temporal analysis ~\cite{Ahn2013EfficientFS,Ibarria2003DynapackSC,Luo2013CompressionO3} predict vertex trajectories for forming vertex groups. There is by far only a handful of works ~\cite{Han2007TimeVaryingMC,Yamasaki2010PatchbasedCF} focusing directly on compressing uncorrelated mesh sequences, i.e., meshes without correspondences. The quality of these approach fall short compared with the ones with correspondences. Our learning based approach in contrast can robustly and accurately process uncorrelated mesh sequences and is able to achieve very high compression rate. Further, we employ the smoothness of the correspondence trajectory as well as geodesic consistencies for fine tuning our solution.

%\section{Unstructure Mesh Compression}
%Our unstructured mesh sequence compression pipeline consists of three main steps. Firstly, we use the panaroma depth image as input to trainned a neaural network for computing dense correspondence between shapes, we revised networks loss function by imposing distance constrain through each feature vector. Then we proposed a temporal geodesic distance fine-tuning step to make correspondence more accurate across sequence and output a animation sequence. Finally, we efficiently compress the animation sequence output from second step using our auto-encoder network.

\section{Vertex Correspondence Matching}
We first present a learning-based scheme to establish vertex correspondences between a pair of models. Since pose changes can greatly affect the appearance of the model, previous approaches train on a very densely sampled viewpoints in the hope that some viewpoints will match the observed ones. We instead introduce a new panoramic depth map (PDM) for efficient training.

\subsection{PDM Generation}
A PDM, in essence, is a panoramic depth map of an inward-looking concentric mosaics (CM)~\cite{Shum1999RenderingWC}, as shown in figure~\ref{fig:PDM_fig}. Our key observation is that each CM covers the complete longitudinal views towards an object, capturing its omni-directional appearance. Traditionally, a CM can be synthesized by first rendering a dense sequence of images on a circle facing towards the human shape and then composing columns at identical locations (e.g., the middle column) from all images. Such a rendering is computationally expensive as it requires rendering a very large number of images.

We instead adopt a GPU-based multi-perspective rendering technique~\cite{Yu2009TowardsMR}: on the vertex shader, we map each vertex of the mesh onto the pixel using the CM projection model while by passing the traditional perspective projection. We also record the depth of the vertex by computing the distance between the vertex and the cylindrical image plane. The rasterization pipeline then automatically generates the PDM via interpolation, and with z-buffer enabled, we resolve the visibility issue when multiple triangles cover the same pixel. Our GPU algorithm is significantly more efficient than the composing approach. In our implementation, we render 6 PDMs at different latitudes, all viewing towards the center of the human object, at \textcolor{black}{20,30,40,50 and 60 degrees respectively}. We find they are sufficiently robust for handling complex occlusions across a variety of poses.

Compared with prior art \cite{wei2016dense} that renders dense perspective depth maps, we use a much smaller images (6 vs. 144) where each PDM provides an omni-directional view towards the human object. More importantly, the PDM better handles occlusions than a regular perspective image. For example, in perspective depth maps, visible shape parts such as the head and the outer surface of the arms and legs appear many more times than "hidden" parts such as the inner parts of the arms and legs, causing not only redundancies but also bias on training. In contrast, in each PDM, each vertex will appear at most once, providing a more reliable training set for extracting shape descriptors.

\subsection{Correspondence Matching}
Next, we train a deep network on the PDMs for computing dense vertex correspondence over temporal frames of the human shape. We formulate the problem as follows: \textcolor{black}{given two sets of vertices on the reference and the target meshes $\mathcal{S} = \{s|s\in\mathbb{R}^3\},\mathcal{T} = \{\tau|\tau\in\mathbb{R}^3\} $}, we set out to find a dense mapping between vertices. Clearly, this mapping should be able to differentiate body parts in order to establish reliable correspondences. We formulate the correspondence matching problem as a classification task: our goal is to \textcolor{black} {train a feature descriptor $f:I_{PDM} \rightarrow \mathbb{R}^d$, which maps each pixel in every input PDM to a feature vector.} For a pair of vertices that belong to the same or nearby anatomical parts, the difference between their feature vectors should be relatively small. For the ones that different anatomical parts, the difference should be large, especially for parts that lie far away from each other.

We construct a network with two modules: the feature descriptor module and the classification module.Fig.~\ref{fig:CorresNetworkPipeline} shows our network architecture.
We indirectly train $f$ with the help of the classification module. To enforce smoothness of the feature descriptor, we partition the classification module into multiple segmentation tasks, one classifier per segmentation. Each classifier aims to assign every vertex and its correspondences across the mesh sequences with an identical label.

%Each classifier's goal is to assign each pixel with label according to different segmentations, and assign same label in different view and same segmentation}. As two points separate farther apart, the possibility of them having the same label should decrease.

We train the feature descriptor $f$ and classifiers simultaneously by adopting the loss function $L_{total} = L_{data} +L_{reg}$, where the data term $L_{data}$ aims to resolves classification as
\vspace{-2em}

\begin{equation}
 \label{eq:data}
\boldsymbol{L_{data}} = -\frac{1}{\mathfrak{N}}\sum_{i}^{\mathfrak{N}} \sum_{p\in I_i}\sum_{j}^K[\mathbbm{1}\{I_{i,p}^{seg}==j\}log(\phi_{j}^{m(i)}(f(p)))]
\end{equation}
\vspace{-1em}
where
\begin{equation}
 \label{eq:data2}
\phi_{j}^{m(i)}(x) = \frac{\exp(\theta_{m(i),j}^T x)}{\sum_l^K\exp(\theta_{m(i),l}^T x)}
\end{equation}

$\mathfrak{N}$ corresponds to training batch size, $i$ refers to the index of a training sample
within a batch, $I_i$ corresponds to the $i_{th}$ input PDM image in current training batch, $I_{i,p}^{seg}$ is the label of pixel $p$ in the $i_{th}$ sample of training batch, $K$ is the number of labels and $M$ is the number of segmentations. $\theta_{m(i),l}$ refers to the parameters of classifier for segmentation $m(i)$. Eq.\ref{eq:data2} can be viewed as an extended Softmax regression model.  $\mathbbm{1}\{\cdot\}$ is the indicator function, so that $\mathbbm{1}\{$a \space true statement$\} = 1$, and $\mathbbm{1}\{$a false statement$\} = 0$.

%$n = \{n_{descriptor},\{n_{classifier}^{i}\}_{i=1}^M\}$
The regularization term $L_{reg}$ aims to make the feature descriptor more distinctive over different anatomical parts as:
\vspace{-1em}
\begin{equation}
    \label{eq:L_reg}
    %\boldsymbol{L_{reg}} = -\sum_{l_i\in\l} ||avg(f(l_i)) - avg(f(\hat{l_i}))||^2
    \boldsymbol{L_{reg}} = -\sum_{i}^\mathfrak{N}\sum_{\mu}\sum_{\nu>\mu} ||avg(f(I_i^{l_\mu})) - avg(f(I_i^{l_\nu}))||^2
     %\boldsymbol{L_{reg}} = -\sum_{\mu=1}^{l_\mu\in\l}\sum_{\nu> \mu}^{l_\nu\in\l} ||avg(f(I_{l_\mu})) - avg(f(l_{l_\nu}))||^2
\end{equation}

\textcolor{black}{where $l_{\mu}$ and $l_{\nu}$ are the $\mu_{th}$ and $\nu_{th}$ label's mask, respectively. $f(I_i^{l_\mu})$ and $f(I_i^{l_\nu})$ are the feature vectors labeled with $\mu$ and $\nu$. $avg(\cdot)$ calculates the average over a sets of vectors.}

%Where $l_i$ is the group of two dimensional index of each PDM belong to labels $i$, f is current panorama depth map's feature descriptor $f = n_{descriptor}(w)$,  .

While several recent approaches have adopted the Alexnet~\cite{Alexa2000RepresentingAB} as the feature descriptor module, we recognize that our problem resembles more the human pose estimation problem where hourglass ~\cite{newell2016stacked} has shown superb performance. Its network structure is capable of downsampling feature maps at multiple scales where it can process each scale via convolutional and batch normalization layers. In our implementation, we conduct nearest neighbor upsampling on feature maps to match across different scales. After each upsampling, a residual connection transmits the information from the current scale (level) to the upsampled scale via element-wise addition. Finally, feature vectors can be extracted across the scales in a single pipeline with the skip layers. \textcolor{black}{We also remove the first max pooling layer to match output resolution with the input.}

\begin{algorithm}

\caption{Improve Correspondence Matching Via Voting}\label{euclid}
\begin{algorithmic}[1]
\label{tab:algorithm}
\Procedure{$Voting(\mathcal{S},\mathcal{T},f_s,f_t,I_s,I_t)$}{}
%\State \FORALL{<condition>} \STATE{<text>}
%	   \ENDFOR
\State $\textit{Initial vote matrix $m_{vote} = zeros(len(f_s^p),len(f_t^q))$}$
\For{\texttt{each source view $p$}}
%\For{\texttt{$j\in{1,...n}$}}
\For{\texttt{each target view $q$}}
\State $\textit{$P_s^p$ = Reproject($I_s^p$)}$
\State $\textit{$P_t^q$ = Reproject($I_t^q$)}$

\State $\textit{$index_s^p = $nnsearch($\mathcal{S}$,$P_s^p$)}$
\State $\textit{$index_t^q = $nnsearch($\mathcal{T}$,$P_t^q$)}$
\State $\textit{$\mathcal{F}_{index} = $nnsearch($f_t^q$,$f_s^p)$}$
\For{\texttt{$k = 1:len(f_s^p)$}}
\State $\textit{$vote_s=index_s^p(k)$}$
\State $\textit{$vote_t=index_t^q(\mathcal{F}_{index}(k))$}$
\State $\textit{$m_{vote}(vote_s,vote_t) \mathrel{+}=1 $}$
\EndFor
%ENDFOR
\EndFor
\EndFor
\For{\texttt{each $row$ in $m_{vote}$}}
\State $\textit{CorresIdx(row) = $argmax_j(m_{vote}(row,j))$}$
\EndFor
%\State $\textit{$CorresIdx = FindMaxIndex(m \textunderscore vote,'row')$)}$
\State \Return $\textit{$CorresIdx$}$

%\State $\textit{stringlen} \gets \text{length of }\textit{string}$
%\State $i \gets \textit{patlen}$
%\BState \emph{top}:
%\If {$i > \textit{stringlen}$} \Return false
%\EndIf
%\State $j \gets \textit{patlen}$
%\BState \emph{loop}:
%\If {$\textit{string}(i) = \textit{path}(j)$}
%\State $j \gets j-1$.
%\State $i \gets i-1$.
%\State \textbf{goto} \emph{loop}.
%\State \textbf{close};
%\EndIf
%\State $i \gets i+\max(\textit{delta}_1(\textit{string}(i)),\textit{delta}_2(j))$.
%\State \textbf{goto} \emph{top}.
\EndProcedure
\end{algorithmic}
\end{algorithm}

Recall that previous methods average the per-pixel feature vectors to obtain a per-vertex feature vector. We adopt a different scheme: assume shape $\mathcal{S}$ and $\mathcal{T}$ have sets of feature vectors $f_s$ and $f_t$ respectively, where $f_{s}^{p}$ and $f_{t}^{q}$ correspond to feature vectors in shape $\mathcal{S}$,$\mathcal{T}$, and the view $p$,$q$ of the 3D model; we build a $len(f_s^p)\times len(f_t^q)$ voting matrix that matches correspondences from the source view $p$'s feature vectors $f_s^p$ to the target view $q$'s feature vectors $f_t^q$ via nearest neighbor search in feature space. We then accumulate the votes into a voting matrix. Finally, we extract the maximum vote index of each row as final correspondence $CorresIdx$. An outline of the algorithm is shown in Algorithm 1, \textcolor{black}{ where $Reproject$ projects PDM $I_s$ and $I_t$ onto 3D points $P^p_s$ and $P^q_t$, $index= nnsearch(X,Y)$ finds the nearest neighbor in $X$ for each point in $Y$, each row in $index$ corresponds to the index of nearest neighbor in $X$ of the corresponding row in $Y$.}

%this paragraph is JUNK! I THINK THE MATHS IS WRONG!!! PLEASE READ AND REVISE.

%We conduct our voting algorithm When computing the correspondence between two shapes $S,T$ with $m$ and $n$ vertices respectively. We construct a $m\times n$ voting matrix. For each source panorama-view depth map per-pixel feature descriptor, we establish correspondence by conduct nearest neighbor search on each target panorama-view depth map feature descriptor. The detail algorithm is illustrate below:
\subsection{Implementation}
For training, we collect training data from the MIT dataset~\cite{Vlasic2008ArticulatedMA}, SCAPE~\cite{Anguelov2005SCAPESC} and Yobi3D ~\cite{wei2016dense}. The MIT dataset contains 10 human objects with the ground truth dense correspondences and we use 7 out of 10 for training (samba,march1,squat1,squat2,bouncing,crane,march1) and the rest 3 (swing, jumping and march2) for testing. SCAPE models a single human subject of $71$ different poses where the poses are registered to form dense correspondences. Yobi3D~\cite{wei2016dense} consists of 2,000 avatars in varies poses. %Yobi3D, however, it does not contain point-to-point dense correspondences. Yet, each model still consists of 33 key points that are manually annotated.
\vspace{-0.2em}
To generate segmentation patches, for the MIT dataset and SCAPE we follow the same strategy as ~\cite{wei2016dense} by segmenting each model into 500 patches. For each mesh sequence with the ground truth correspondences, \textcolor{black}{we generate each segmentation by randomly select 10 points on each model. We then add the remaining points using farthest point-sampling and obtain the segmentation by using those sample points as cluster center. Finally, we propagate the initial segmentations onto consecutive frames using the known dense correspondences. For yobi3D data, recall that no dense correspondences are available across the models, we use manually semantic annotated key points as the cluster centers to generate segmentations.}

%we first randomly select 10 points in the model and then generate the rest via farthest point sampling (XXX what do you mean? specify). We then propagate the initial segmentations onto consecutive frames using the known dense correspondences. Overall, each model is segmented into 30 components.
%Dense correspondence network consists of two modules,
%feature extraction module and classification module.
Our feature extraction module takes the PDMs as input and is trained on 2-cascaded-level hourglass network with the first max-pooling layer removed. In each classification layer, we use a convolution layer with $1\times1$ filter size to replace fully-connected layer and conduct 2D Softmax operation on feature vectors generated by descriptor. Different classifier is trained for different body parts across all frames in the mesh sequence but the descriptor's parameters remain the same. Our network is rather large on high resolution PDMs($512\times512$ resolution) and we handle this by training a batch size of 4. The training time is approximately 48 hours on a single Titan X Pascal GPU.
%Our network is rather large on high resolution feature maps and we handle this by training a batch size of 4.
\begin{figure}
\centering
\includegraphics[width=0.9\linewidth]{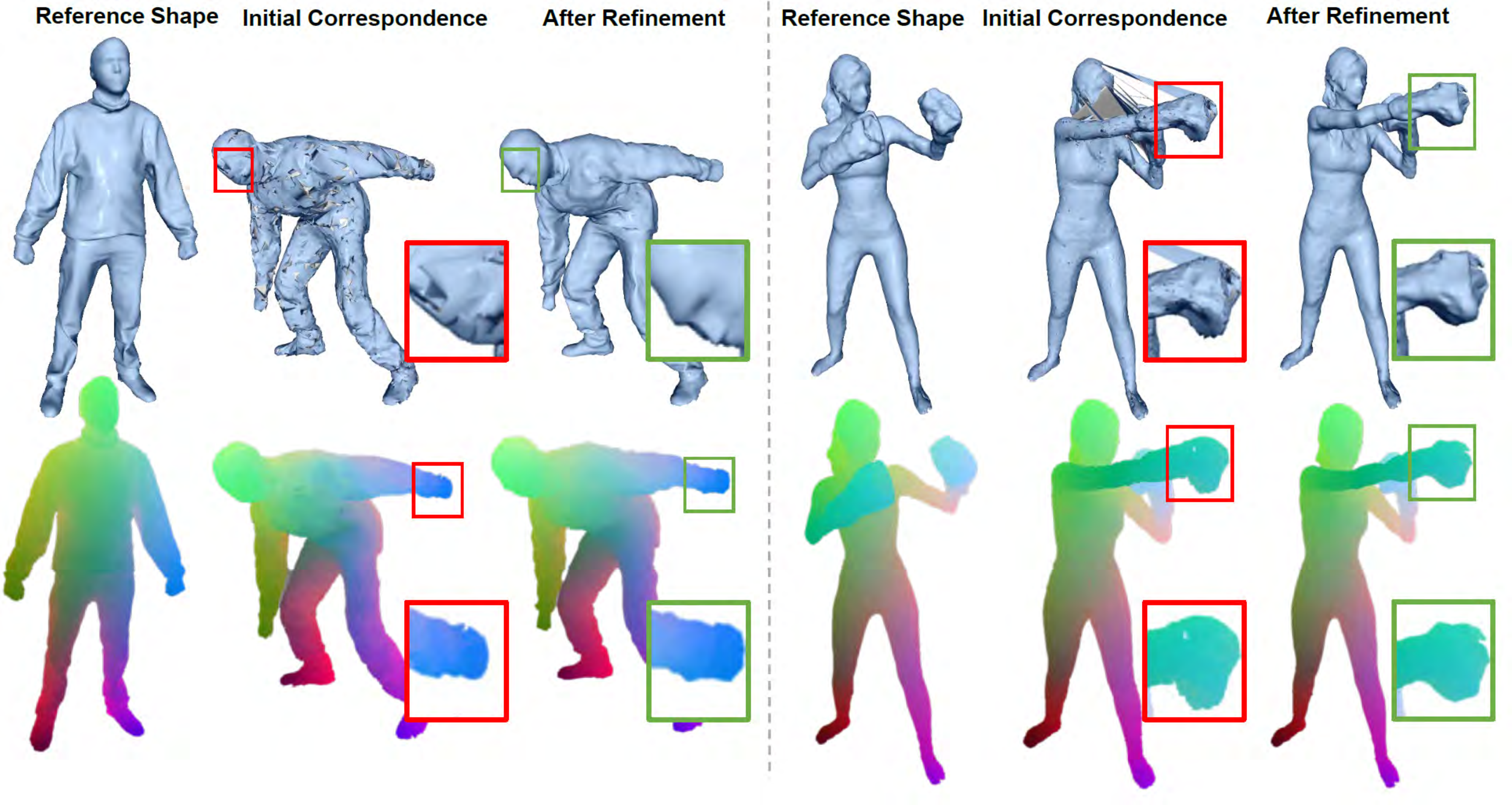}
\caption{Result before and after refinements in the Jumping and Boxing sequences. For each sequence, from left to right we show the reference mesh, the target mesh with initial correspondence matching, the target mesh after refinement. The top and bottom rows show the geometry and correspondence maps respectively.}
\label{fig:fineTuning} %temporal_hint
\end{figure}
%$\{P^N|_{n=1,..,N}\}$
\vspace{-0.5em}
\section{Correspondence Refinement}
Our correspondence matching scheme processes any two consecutive frames of meshes. We further refine the results to handle a $N$-frame motion sequence $\{P_n\}_{n=1}^N$ where dense correspondences are maintained coherently across the frames. The challenges in processing real data is that they contain noise and topological inconsistencies. We therefore first locate a reference frame $P_{\kappa}$ that has the lowest genus. Assume the mesh contains $V$ vertices, to reduce drifting we compute correspondences $\{c^n_i\}_{i=1}^V$ between $P_k$ and every other frame $P_n$ using our feature descriptor. As a result, we obtain a vertex trajectory matrix $A$ that stores each frame in its columns:

\begin{equation}
    \label{eq:vertex trajectory}
    %\boldsymbol{L_{reg}} = -\sum_{l_i\in\l} ||avg(f(l_i)) - avg(f(\hat{l_i}))||^2
     \boldsymbol{A} = \begin{pmatrix}
     					c_1^1 & c_1^2 & ... & c_1^N \\
                        c_2^1 & c_2^2 & ... & c_2^N \\
                        ...   & ...   & ... &  ... \\
                        c_V^1 & c_V^2 & ... & c_V^N\\
     				  \end{pmatrix} =
                      \begin{pmatrix}
                      T_1 \\
                      T_2 \\
                      ...\\
                      T_V\\
                      \end{pmatrix}
\end{equation}
where $T_i$ is a row vector that represents each vertex correspondence trajectory $\{c_i^n\}_{n=1}^N$.

Figure~\ref{fig:fineTuning} illustrates that, even majority of the correspondences are accurate, a small percentage of outliers can cause severe distortions on mesh surfaces. We therefore conduct a correspondence refinement step on the vertex matrix $A$ using geodesic and temporal constraints similar to ~\cite{huang2008non}. By assuming the deformation is isometric, we first find the correspondence outliers $c^n_i$ in each vertex trajectory $T_i$ based on the geodesic distance consistency measurement. We then refine each outliers by imposing temporal smoothness $E_{temporal}$ and geodesic consistency $E_{geodesic}$ constraints as:
\begin{equation}
    \label{eq:fine tuning temporal}
%\boldsymbol{L_{total}} = \sum_{i=1}^M n(w,w_i^c,I,I_i^{seg}) + %\boldsymbol{L_{reg}}
	\boldsymbol{E_{temporal}} = ||c_i^{(n+1)} + c_i^{(n-1)} -2t||^2
\end{equation}
and
\vspace{-1em}
\begin{multline}
    \label{eq:fine tuning geodesic}
%\boldsymbol{L_{total}} = \sum_{i=1}^M n(w,w_i^c,I,I_i^{seg}) + %\boldsymbol{L_{reg}}
	\boldsymbol{E_{geodesic}} = \sum_{(c^n_k,c^{n\pm1}_{k})\in\Theta} [d_g(c^{n-1}_i,c^{n-1}_k) - d_g(t,c^{n}_k)]^2\\
    + [d_g(t,c^n_k) - d_g(c^{n+1}_i,c^{n+1}_k)]^2
\end{multline}
where $d_g$ is the geodesic distance between two points, $\Theta$ is the set of confident correspondence set among three frames $n-1,n,n+1$. To enforce geodesic consistency, given a trajectory outlier $c^n_i$ in vertex trajectory $i$, we find the nearest $c^n_j$ that has a highly confident correspondence to $c^{n-1}_j$ and $c^{n+1}_j$, and assign $c^n_j$ as the adjusted position $t$ to replace $c^n_i$. Next, we construct the geodesic term $E_{geodesic}$ to enforce the geodesic distance between each pair of correspondence in frame $n$  close to each pair of correspondences in frame $n-1$ and $n+1$. To enforce temporal smoothness, we utilize a temporal term $E_{temporal}$ by assuming each the outlier vertex $c_i^n$ in frame $n$ should close to the middle of the two adjacent corresponding vertex $c_i^{n+1}$ and $c_i^{n-1}$ respectively. To find an optimal $t$ to minimize $E_{refine} = E_{temporal} + E_{geodesic}$, we set out to refine each individual correspondence outlier $c^n_i$ by searching the $\mathcal{K}$ nearest neighbor of $t\in\Omega(c^n_j)$. Figure~\ref{fig:fineTuning} compares the results before and after the refinement: the reconstructed mesh surface contains much fewer artifacts after refinement.

\begin{figure}
\centering
\includegraphics[width=0.9\linewidth]{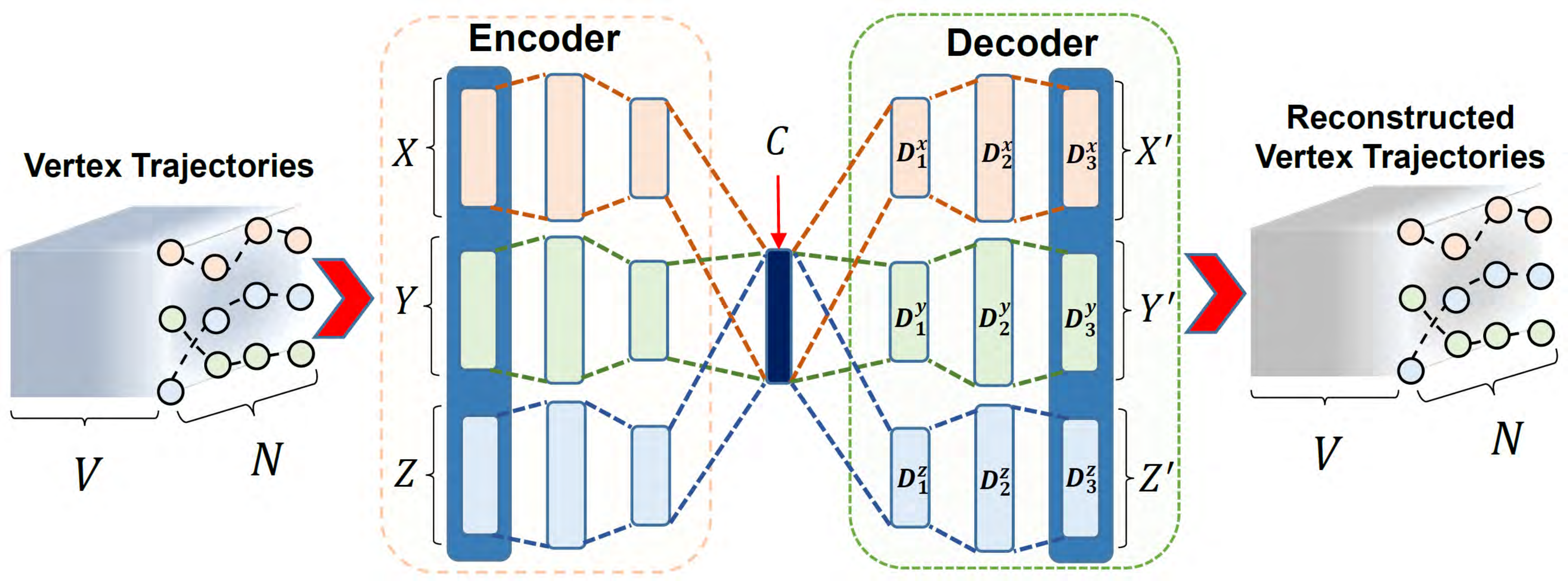}
\caption{Our Autoencoder network for mesh compression/decompression. The x, y, and z dimensions of the vertex trajectories are separately processed. Each block, except the output one, represents a fully-connected layer with ReLU as its active function. }
\label{fig:AutoEncoder} %temporal_hint
\end{figure}

%compress each trajectory $T_n$ in
%(XXX? are you sure this is the correct term?)
\section{Mesh Compression and Decompression}
With the refined vertex correspondences, we can effectively convert the input mesh sequence to an animation mesh with a consistent topology. Recall that an animation sequence should have consistent vertices and connectivities, we simply need to compress vertex trajectories $A=\{T_i\}_{i=1}^V$.

Traditional dimension reduction techniques are commonly used for compressing animation meshes. We adopt the Autoencoder framework, as shown in Figure~\ref{fig:AutoEncoder}. we present a 7-layer parallel Autoencoder network structure with 3D vertex trajectories as input. In the encoder path, to encode the trajectory according to $x,y,z$ coordinate separately, we split $\{T_i\}_{i=1}^V$ into three parts: $\{T_i^x\}_{i=1}^V$,$\{T_i^y\}_{i=1}^V$ and $\{T_i^z\}_{i=1}^V$, and feed the above three parts into three parallel networks respectively. The three parallel network will then merge into an intermediate layer $C$ which is a compressed representation of the input data.

The decoder path is the inverse operation of encoder. For decompression, we efficiently extract the trained parameters from the intermediate layer $C$ and the rest of decoder layer $D_1^{x,y,z},D_2^{x,y,z},D_3^{x,y,z}$ to conduct a forward operation to reconstruct the entire animation sequence. In our training process, we construct layers with varying sizes to achieve different bpvf (bit per frame per vertex). Our training process uses a batch size of 200, and the process converges after about 6,000 iterations on the GPU. Compared with traditional Principle Component Analysis (PCA) based approaches, our solution supports nonlinear trajectories and therefore is much more effective in both compression rate and quality, as shown later in the experiment.

%\begin{equation}
%    \label{eq:mesh decode formula}
%\boldsymbol{L_{total}} = \sum_{i=1}^M n(w,w_i^c,I,I_i^{seg}) + %\boldsymbol{L_{reg}}%
%	\boldsymbol{E_{temporal}} = ||c_i^{(n+1)} + c_i^{(n-1)} -2t||^2
%\end{equation}

%Mesh trajectory network training detail is as follows. The input is %each vertex trajectory in a $N\times3$vector order with $x$,$y$,$z$ %respectively, where $N$ is number of frame in each sequence. While %middle layer is the compress data representation. The encoder layer %is fix, the first layer size XXX, second layer size XXXXX. We use %different size of the decode layer in order to achieve different %bpvf(bit per frame per vertex). we training this unsupervised %network with batch size equal to the mesh vertex number, and it %takes roughly 6000 iteration upon converge.

%So we have to
%\section{Implementation Details}
%We first describe the detail in supervised network we used to generate dense correspondence, then illustrate the detail the mesh sequence compression unsupervised network.
%\subsection{Mesh Trajectory Compression Network}
\begin{figure}
\centering
\includegraphics[width=0.9\linewidth]{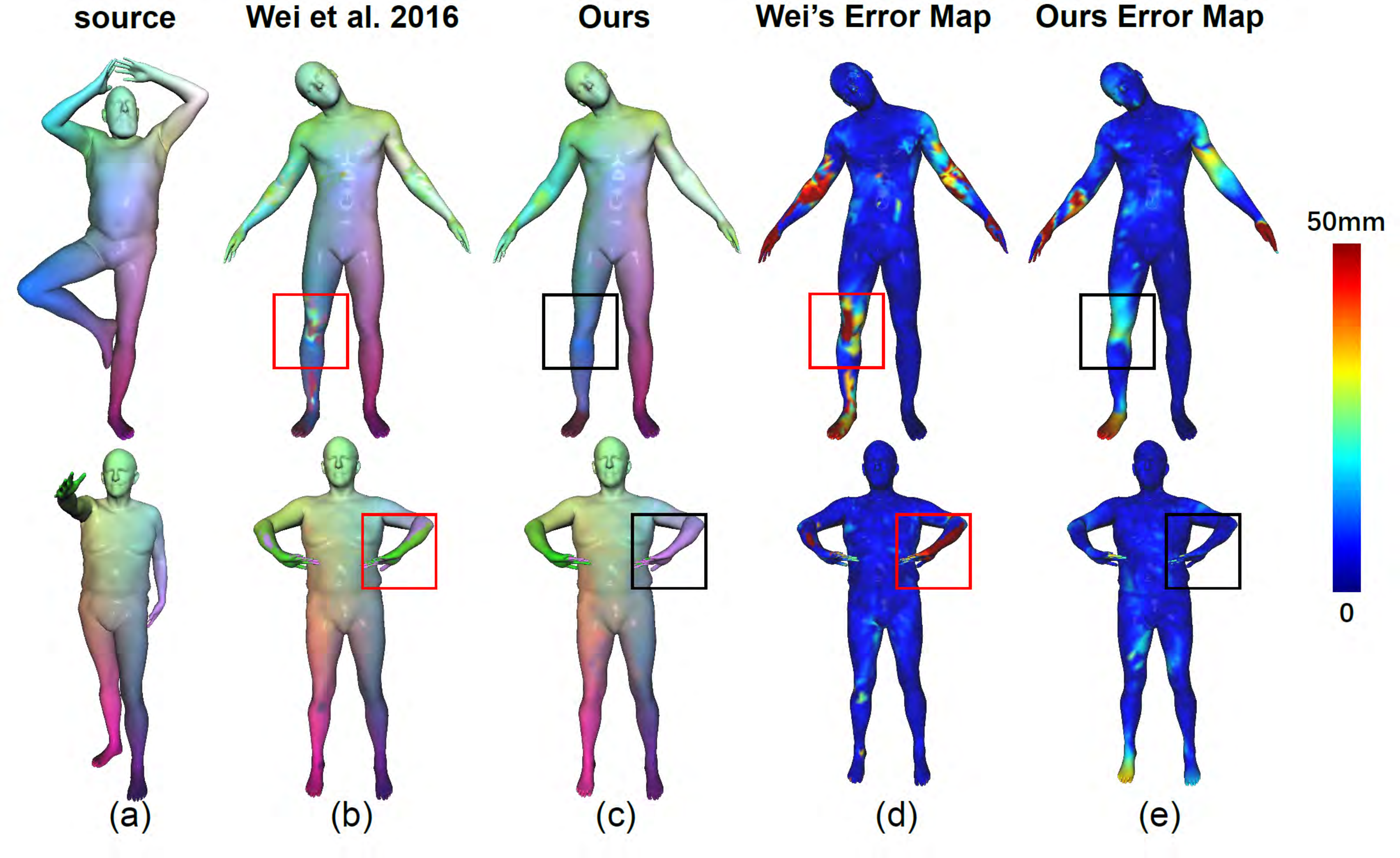}
\caption{Comparisons on FAUST~\cite{bogo2014faust}. (a) shows the reference mesh; (b) and (c) show the results by~\cite{wei2016dense} and ours; (d) and (e) show the corresponding errors maps. Our technique more robustly handles strong deformations (e.g., knee and elbow bending).}
\label{fig:FasustResult} %temporal_hint
\end{figure}

\begin{figure}
\centering
\includegraphics[width=0.9\linewidth]{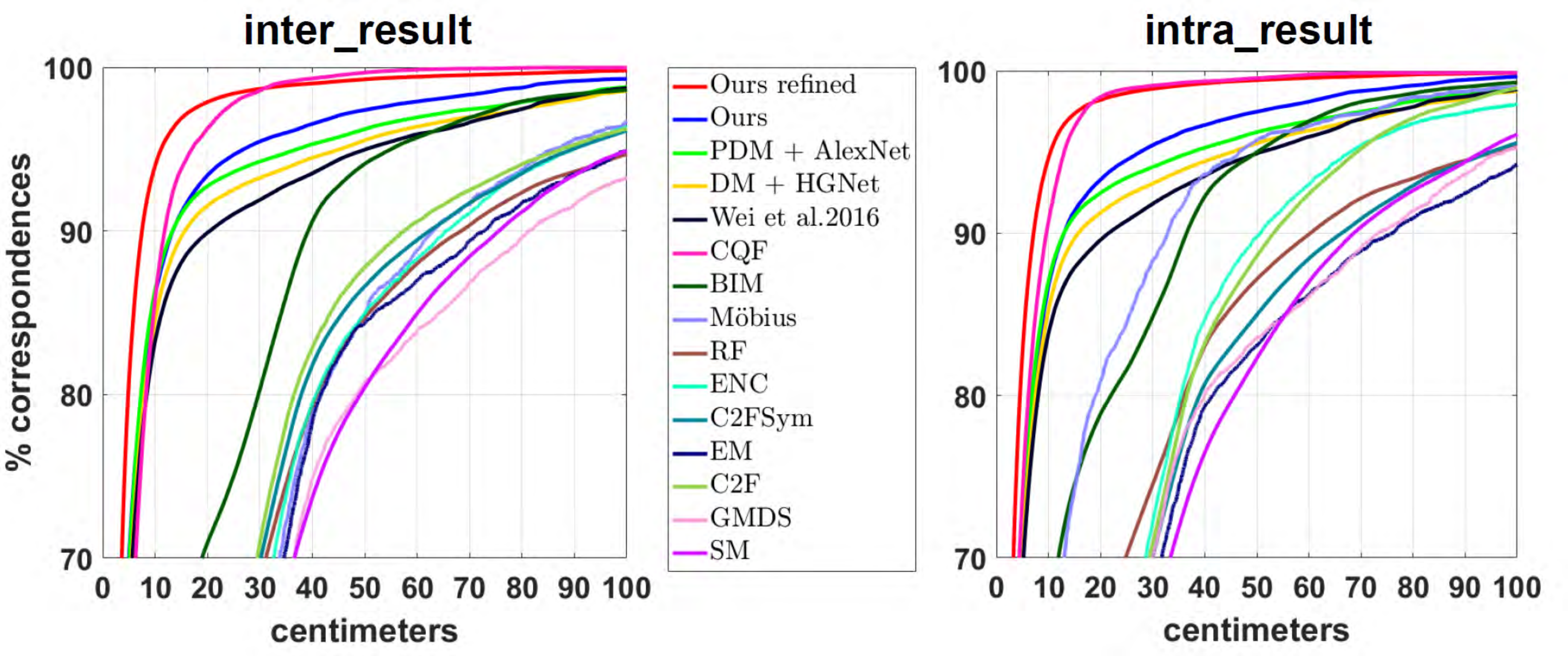}
\caption{Quantitative evaluations on FAUST~\cite{bogo2014faust}. Our technique outperforms the state-of-the-art~\cite{wei2016dense}, especially after the refinement process.}
\label{fig:FaustResultPlot} %temporal_hint
\end{figure}
%Guo2015RobustNM

\begin{figure*}
\centering
\includegraphics[width=0.8\linewidth]{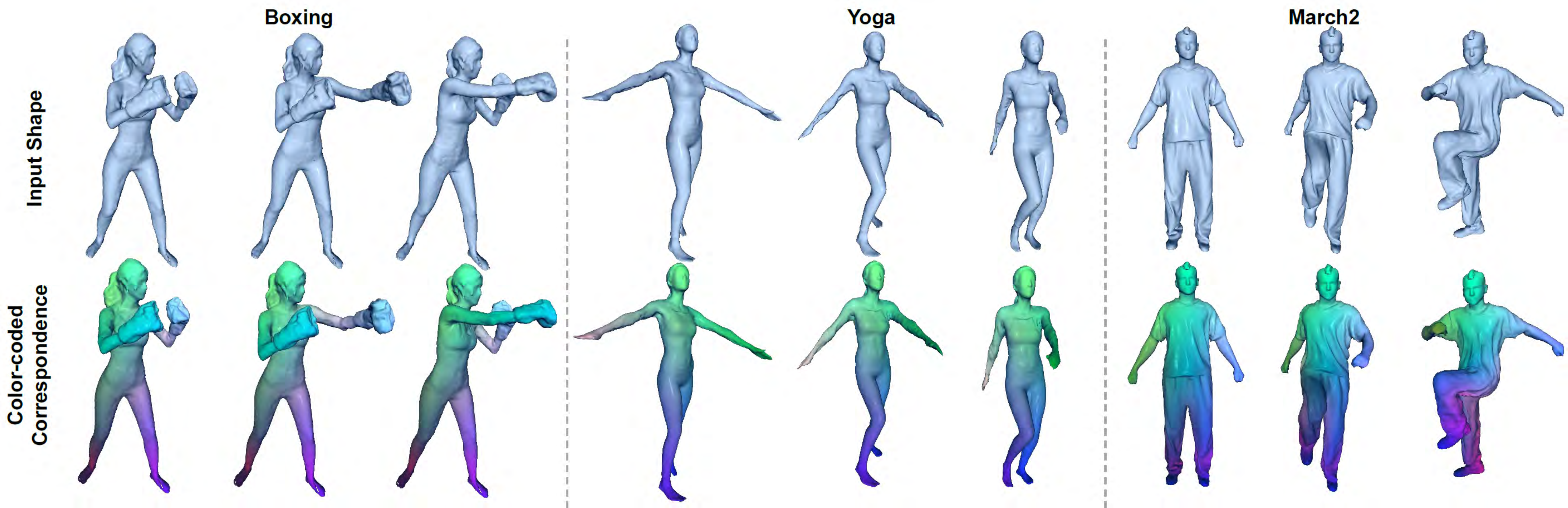}
\caption{Results using our technique on the Boxing, Yoga datasets and MIT~\cite{vlasic2009dynamic} March2 datasets.}
\label{fig:refinement3Result} %temporal_hint
\end{figure*}

\section{Experimental Results}
We conduct comprehensive experiments on both publicly available datasets ~\cite{vlasic2009dynamic,bogo2014faust} and our own 4D human body dataset. Our own capture system is composed of 32 synchronized industry cameras at a 720P resolution and 24 fps. We then apply structure-from-motion for camera pose calibration and subsequently use open-source multi-view stereo matching solutions ~\cite{Furukawa2007AccurateDA} for generating the point cloud. To construct the mesh from the point cloud, we apply Poisson surface reconstruction ~\cite{Kazhdan2006PoissonSR}. Notice that the initially reconstructed meshes do not have vertex correspondences. In the paper, we demonstrate 3 full body motion sequences:\textcolor{black}{ Yoga,Wakingup and Boxing has 360,300,270 frame respectively, each data with average 32K vertex number per mesh.}
%XXX raw datasize, how many frames, how many vertices each frame, what are the texture sizes, what are the total sizes.

Our pipeline conducts the three-step approach to simultaneously compute the dense correspondences across the frames and then use the results to compress the mesh sequence. All experiments (including training and testing) are performed off-line on a PC with CPU Intel Core i7-5820K, 32 GB memory and a Titan X GPU.\textcolor{black}{ On the computation overhead, the average cost for mesh correspondence generation process is 21 secs for establishing the initial dense correspondences and 11 secs for correspondence refinement. Our mesh compression step takes about average 64 secs for compressing each entire sequence and 5 secs for decompression.}
%XXX are the length the same for all sequences?? otherwise the number does not make sense) and 5 secs for decompression.}
\vspace{-1em}
\paragraph{Correspondence Matching Results.}  To further demonstrate the effectiveness of our learning-based correspondence matching technique, we experiment on the FAUST dataset~\cite{bogo2014faust}. FAUST is a public dataset composed of training and testing data. The training data has ground truth dense correspondences across the frames but the testing data does not. We conduct the experiment on the training dataset that has 100 shapes, includes 10 human subjects with 10 different poses. We have conducted two different types of evaluation. The first computes the correspondence between inter-object: source and target are of different human subjects with the different poses and the second between intra-object: source and target are of the same human object but with the different poses. The results shown in Fig.~\ref{fig:FasustResult} demonstrate our method incurs less error compared to the state-of-the-art ~\cite{wei2016dense} based on our own implementation. In Fig.~\ref{fig:FaustResultPlot}, we conduct quantitative evaluations and show error distributions in centimeters. Other techniques including GMDS~\cite{bronstein2006generalized}, Mobius voting~\cite{lipman2009mobius}, blended intrinsic maps (BIM)~\cite{Kim2011BlendedIM}, coarse-to-fine matching (C2F) \cite{sahillioǧlu2011coarse}, the EM algorithm~\cite{sahilliouglu2012minimum}, coarse-to-fine matching with symmetric flips (C2FSym)~\cite{sahilliouglu2013coarse}, sparse modeling (SM) ~\cite{Pokrass2013SparseMO}, elastic net constraints (ENC)~\cite{Rodol2013ElasticNC}, and random forests (RF)~\cite{Rodol2014DenseNS} were based on the implementations by Chen et al.~\cite{chen2015robust}. Figure.~\ref{fig:FaustResultPlot} shows comparisons on accuracy of our technique vs. others on all intra-subject pairs and all inter-subject pairs. We also conduct self-evaluation to compare the modified hourglass architecture
vs. ~\cite{wei2016dense} with the same PDM inputs. As shown in Figure.~\ref{fig:FaustResultPlot}, the hourglass architecture outperforms ~\cite{wei2016dense} using either traditional depth maps or PDMs as inputs while PDMs still significantly outperform the regular depth maps.
%We will include this discussion.

%Where comparison method include GMDS~\cite{bronstein2006generalized}, Mobius voting~\cite{lipman2009mobius}, blended intrinsic maps (BIM)~\cite{Kim2011BlendedIM}, coarse-to-fine matching (C2F) \cite{sahillioǧlu2011coarse}, the EM algorithm~\cite{sahilliouglu2012minimum}, coarse-to-fine matching with symmetric flips (C2FSym)~\cite{sahilliouglu2013coarse}, sparse modeling (SM) ~\cite{Pokrass2013SparseMO}, elastic net constraints (ENC)~\cite{Rodol2013ElasticNC}, and random forests (RF)~\cite{Rodol2014DenseNS}. Each method is evaluated on all intra-subject pairs and all inter-subject pairs as shown in Figure.~\ref{fig:FaustResultPlot.}

%Where comparison method include GMDS~\cite{bronstein2006generalized}, Mobius voting~\cite{}, blended intrinsic maps (BIM)~\cite{}, coarse-to-fine matching (C2F) \cite{}, the EM algorithm~\cite{}, coarse-to-fine matching with symmetric flips (C2FSym)~\cite{}, sparse modeling (SM) ~\cite{}, elastic net constraints (ENC)~\cite{}, and random forests (RF)~\cite{}. Each method is evaluated on all intra-subject pairs and all inter-subject pairs as shown in Figure.~\ref{fig:FaustResultPlot.}

\begin{figure}
\centering
\includegraphics[width=0.9\linewidth]{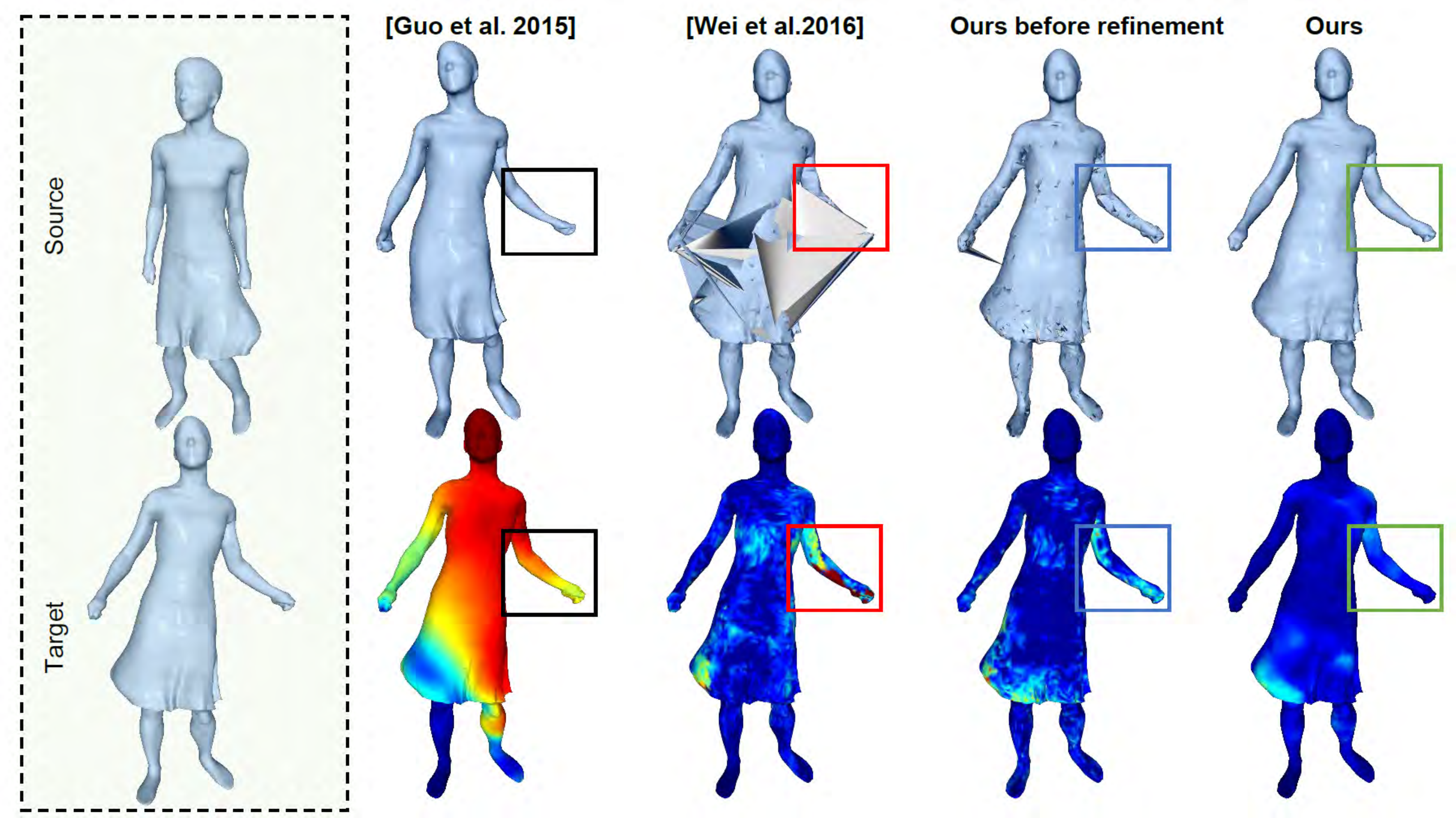}
\caption{Our technique vs. the state-of-the-art on the Swing sequence. ~\cite{Guo2015RobustNM} loses track due to large deformations of the skirt (see geometry inconsistencies); ~\cite{wei2016dense} is able to track most of the vertices but the errors produce topology inconsistencies; Our result before refinement outperforms both in correspondence matching. The results are further improved after refinements. The video results can be found in the supplementary materials.}
\label{fig:refinementResult} %temporal_hint
\end{figure}
%\vspace{-1em}
\begin{figure}
\centering
\includegraphics[width=\linewidth]{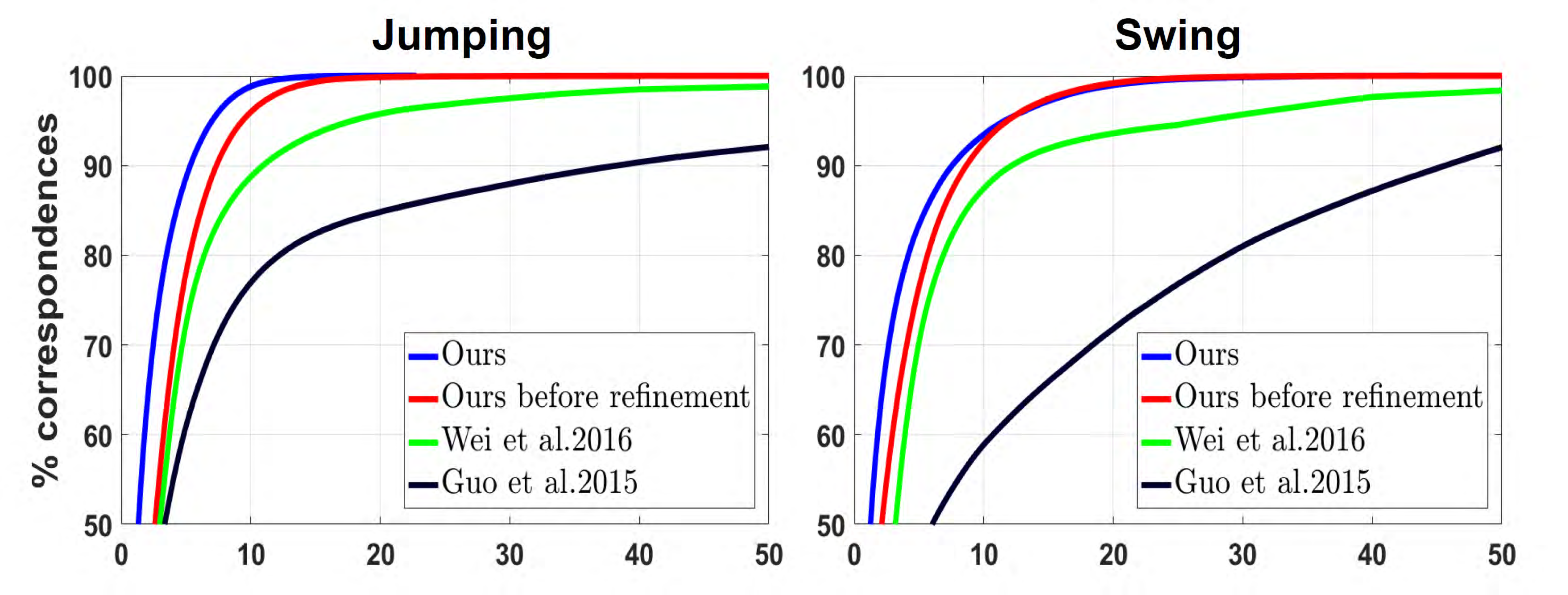}
\caption{Quantitative Comparisons of ours vs. ~\cite{Guo2015RobustNM} and ~\cite{wei2016dense} on highly non-rigid Jumping and Swing sequences.}
\label{fig:Quantitative_refine} %temporal_hint
\end{figure}

\begin{figure}
\centering
\includegraphics[width=\linewidth]{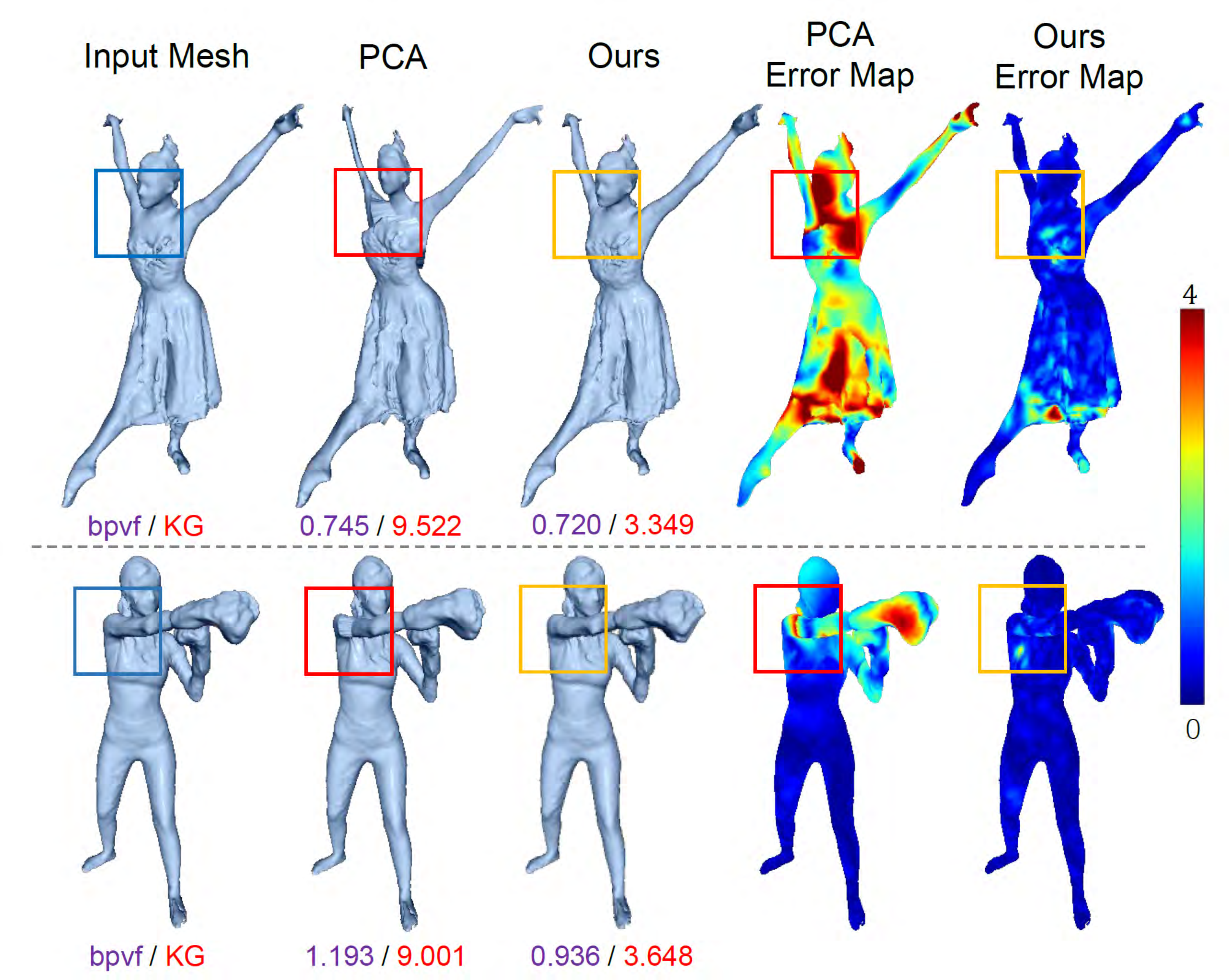}
\caption{Visual comparisons of ours vs. the PCA technique ~\cite{sattler2005simple} on mesh compression. Our technique outperforms ~\cite{sattler2005simple} even with at lower bpvf.}
\label{fig:MCErrorMap} %temporal_hint
\end{figure}
% Starting from the second column,the top row shows the results based on the state-of-the-art non-rigid sequence registration approach~\cite{Guo2015RobustNM}

Next we show how our refinement step further improves the correspondence matching results in Fig. ~\ref{fig:refinementResult}. We only evaluate this process on the MIT dataset, the only one with ground truth dense correspondences. The first column in Fig.~\ref{fig:refinementResult} shows the reference and target meshes from up to bottom. Starting from the second column, we first show the results~\cite{Guo2015RobustNM} among the state-of-the-art non-rigid surface alignment methods~\cite{Newcombe2015DynamicFusionRA,Allain2015AnEV,Budd2012GlobalNA,li2017robust}, previous CNN approach ~\cite{wei2016dense}, our technique before refinement, and ours after refinements respectively; the second row shows the corresponding error maps using various techniques. We observe that due to  large deformations between frames, ~\cite{Guo2015RobustNM} can lose tracking and produce relatively large errors. Our technique before refinement already contains fewer artifacts compared with the previous learning-based approach ~\cite{wei2016dense}. The artifacts were further reduced after refinement. Fig.\ref{fig:Quantitative_refine} shows the quantitative evaluations using our method vs. prior art on the jumping and swing sequences. Fig.~\ref{fig:refinement3Result} shows additional results on Boxing, Yoga and March2.

\begin{figure}
\centering
\includegraphics[width=\linewidth]{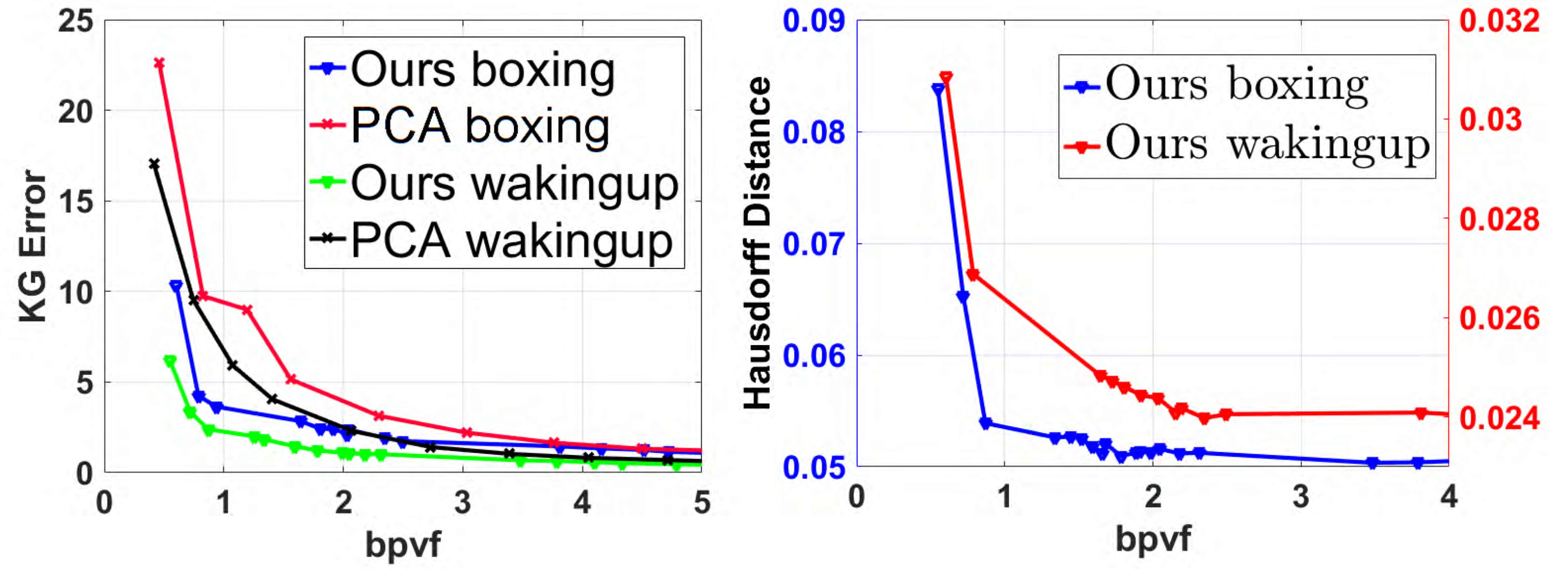}
\caption{Quantitative comparisons of mesh compression on the Boxing and Wakingup sequences using KG error and Hausdorff distance error measures. }
\label{fig:KGerr} %temporal_hint
\end{figure}

%We could see that after enforcing geodesic and temporal constraint, our correspondence result is better in terms of no matter error measurement and smooth of the mesh geometry. More result is shown in fig.~\ref{fig:refinement3Result}
\vspace{-1.5em}
\paragraph{Mesh Compression Results.} Based on the vertex correspondences, we construct animation mesh sequences with consistent connectivity and apply our  Autoencoder neural network efficient data compression. We conduct experiments on six motion sequences (jumping, march2, swing, yoga, boxing, and dance). We first the evaluate the quality degradation caused by compression. Specifically, we measure the distortion errors between the original and the (de)compressed results by using the well-established vertex-based error metric: KG error~\cite{Karni2000SpectralCO}. KG error measures the quality of vertex position reconstruction over an entire sequence as: $\mathbf{D_{KG}} = 100\cdot\dfrac{\|B-\hat{B}\|}{\|B-E(B)\|}$ where $B$ is a matrix representing the original mesh sequence data with size $3V\times N$, $\hat{B}$ is same sized matrix that stands for the mesh sequence after reconstruction. $E(B)$ is a matrix consist of the average vertex position for all frames. Fig.\ref{fig:MCErrorMap} shows the comparisons on our technique vs. Sattler et al.~\cite{sattler2005simple} (our own implementation) at different $bpvf$ (bits per vertex per frame) , we will shown the our bpvf metric in supplementtal material. Fig.~\ref{fig:MCErrorMap} and Fig.~\ref{fig:KGerr} shows that ~\cite{sattler2005simple} introduces relative high distortions and errors when $bpvf$ is low whereas our Autoencoder approach significantly suppresses the errors.
%~\textcolor{red}{$bpvf= \\ \dfrac{6V+q(\mathcal{N}(C)\cdot (V + \mathcal{N}(D_1)) + \sum_{i=1}^2{\mathcal{N}(D_i)\cdot\mathcal{N}(D_{i+1})})}{FV}$,}where $6V$ encodes the connectivity. We use the Edgebreaker~\cite{Rossignac1999EdgebreakerCC} method to compress the connectivity information. $\mathcal{N}()$ define the number of nodes in the corresponding layer. We conduct quantization on each nodes to half float by only using $q= 16$ bits. where our bpvf can be calculated as:

%We further measure the error on the animated mesh sequences before and after compression.

We further measure the error between the original input mesh sequences and mesh sequences after decompression. It is important to note that there is no longer dense correspondences after we decode the compressed mesh sequences. We therefore use the mean Hausdorff distance $\mathcal{H}$ to measure the geometric deviations between the original and the decoded meshes as $\mathbf{D_{Hausdorff}} =\sum_i^N{\mathcal{H}_i}$, where $N$ is the frame number of the sequence, $\mathbf{D_{Hausdorff}}$ computes the average Hausdorff distance of entire sequences at different $bpvf$. Fig.~\ref{fig:KGerr} shows that our decoded mesh sequence has a very low Hausdorff error for over various motion sequences.
%(xxx: are you sure?)
\vspace{-0.5em}
\section{Conclusions and Future Work}
We have presented a learning-based approach for compressing 4D
human body sequences. At the core of our technique is a novel temporal vertex
correspondence matching scheme based on the new representation of
panoramic depth maps or PDM. The idea of PDM is borrowed from earlier
panoramic rendering techniques such as concentric mosaics~\cite{Shum1999RenderingWC} and
multi-perspective rendering ~\cite{Rademacher1998Multiple,Yu2004A} that samples
omni-directionally the appearance of a target object (in our case the
depth map of an human body). By extending existing deep learning
frameworks, our technique manages to learn how to reliably label
vertices into meaningful semantic groups and subsequently establishes
correspondences. We have further developed an autoencoder-based
network that directly uses correspondences for simultaneous texture
and geometry compression. Regarding the limitation, topology changes and occlusions may cause the correspondence tracking failure.
A potential solution is to partition the sequence into shorter, topologically coherent segments.

Alternatively FVVs can be produced via image-based rendering such as view morphing where
new views can be synthesized by interpolating from acquired reference
views without completely obtaining the 3D geometry. Our immediate
future task hence is to extend our approach to handle such cases.
We also plan to experiment on applying our technique for 3D completion: a
partial scan, e.g., a depth map of the model, can be registered onto a
reference, complete model using our technique and missing parts can be
completed via warping.

%i.e., establishing correspondences and conduct compression.

%Finally, we will disseminate our new 4D human body
%data and our solution to the community for further improvements and

%Our work is motivated by emerging free-viewpoint video (FVV). There
%are generally two classes of approaches for generating FVVs, geometric
%reconstruction and image-based rendering. Our current solution aims to
%serve as a compression engine for the former where the geometry is
%obtained via specially designed 3D acquisition systems such as a
%multi-camera dome with or without dense sensors. Alternatively FVVs
%can be produced via image-based rendering such as view morphing where
%new views can be synthesized by interpolating from acquired reference
%views without completely obtaining the 3D geometry. Our immediate
%future task hence is to extend our approach to handle such cases,
%i.e., establishing correspondences and conduct compression. We also
%plan to experiment on applying our technique for 3D completion: a
%partial scan, e.g., a depth map of the model, can be registered onto a
%reference, complete model using our technique and missing parts can be
%completed via warping. Finally, we will disseminate our new 4D human body
%data and our solution to the community for further improvements and
%comparisons.

\section*{Acknowledgement}
This work is partially supported by National Science Fundation under the Grant 
CNS-1513031. Majority of the work was performed while Zhong Li was an intern at 
Plex-VR Inc.

{\small
\bibliographystyle{ieee}
\bibliography{egbib}
}

\end{document}